\documentclass[sigconf,nonacm]{acmart}
\usepackage{graphicx}
\usepackage{amsmath}
\usepackage{algorithm}
\usepackage{algorithmicx}
\usepackage{algpseudocode}
\usepackage{cleveref}
\AtBeginDocument{%
  }

\setcopyright{acmlicensed}
\copyrightyear{2018}
\acmYear{2018}
\acmDOI{XXXXXXX.XXXXXXX}
\acmConference[Conference acronym 'XX]{Make sure to enter the correct
  conference title from your rights confirmation email}{June 03--05,
  2018}{Woodstock, NY}
\acmISBN{978-1-4503-XXXX-X/2018/06}

\begin{document}

\title{Introducing the Swiss Food Knowledge Graph: \texorpdfstring{\\}{ }
AI for Context-Aware Nutrition Recommendation}
\renewcommand{\shorttitle}{Introducing the Swiss Food Knowledge Graph}

\author{Lubnaa Abdur Rahman}
\orcid{0000-0001-7271-2814}
\affiliation{%
\institution{Graduate School for Cellular and Biomedical Sciences}
  \institution{University of Bern}
  \city{Bern}
  \country{Switzerland}
}
\email{lubnaa.abdurrahman@unibe.ch}

\author{Ioannis Papathanail}
\orcid{0000-0002-7767-3831}
\affiliation{%
  \institution{University of Bern}
  \city{Bern}
  \country{Switzerland}
}
\email{ioannis.papathanail@unibe.ch}

\author{Stavroula Mougiakakou}
\orcid{0000-0002-6355-9982}
\affiliation{%
  \institution{University of Bern}
  \city{Bern}
  \country{Switzerland}
}
\email{stavroula.mougiakakou@unibe.ch}

\renewcommand{\shortauthors}{Abdur Rahman L. et al.}

\begin{abstract}
Artificial intelligence has driven significant progress in the nutrition field, especially through multimedia-based automatic dietary assessment. 
However, existing automatic dietary assessment systems often overlook critical non-visual factors, such as recipe-specific ingredient substitutions that can significantly alter nutritional content, and rarely account for individual dietary needs, including allergies, restrictions, cultural practices, and personal preferences.
In Switzerland, while food-related information is available, it remains fragmented, and no centralized repository currently integrates all relevant nutrition-related aspects within a Swiss context.
To bridge this divide, we introduce the Swiss Food Knowledge Graph (SwissFKG), the first resource, to our best knowledge, to ever unite recipes, ingredients, and their substitutions with nutrient data, dietary restrictions, allergen information, and national nutrition guidelines under one graph. 
We establish a Large Language Model (LLM)-powered enrichment pipeline for populating the graph, whereby we further present the first benchmark of four off-the-shelf (<70 B parameter) LLMs for food knowledge augmentation. Our results demonstrate that LLMs can effectively enrich the graph with relevant nutritional information.
Our SwissFKG goes beyond recipe recommendations by offering ingredient-level information such as allergen and dietary restriction information, and guidance aligned with nutritional guidelines. 
Moreover, we implement a Graph-Retrieval Augmented Generation (Graph-RAG) application to showcase how the SwissFKG's rich natural-language data structure can help LLM answer user-specific nutrition queries, and we evaluate LLM-embedding pairings by comparing user-query responses against predefined expected answers. 
As such, our work lays the foundation for the next generation of dietary assessment tools that blend visual, contextual, and cultural dimensions of eating. 
\end{abstract}




\maketitle

\section{Introduction}

Nutrition plays a critical role in addressing pressing health and environmental challenges \cite{papathanail2021evaluation}.
Non-communicable diseases (NCDs) currently account for roughly 75\% of all deaths each year, with cardiovascular diseases (CVDs) being the world’s leading cause of mortality \cite{WHO_NCDs_2024}. 
Diabetes prevalence is projected to climb to 853 million by 2050 \cite{IDF_Atlas_Global_2025}. 
Food allergies are also on the rise: an estimated 4.3\% of the global population lives with food allergies \cite{feng2023prevalence}. 
At the same time, growing health and environmental concerns are driving dietary shifts toward plant-based eating \cite{tziva2020understanding}. 
In Switzerland, these global patterns are reflected locally: CVDs accounted for 27.5\% of all deaths in 2022 \cite{rosemann2024kardiovaskulare}, diabetes affects nearly 500,000 people \cite{swissmedical_diabetology}, and an estimated 4–8\% of the population lives with food allergies \cite{aha_food_allergy_2024}. 
Concurrently, efforts to reduce meat consumption are gaining momentum \cite{zumthurm2025reducing}.

The field of nutrition has witnessed remarkable growth in recent years, with Artificial Intelligence (AI) playing a pivotal role in automating processes such as dietary assessment. 
Image-based automatic dietary assessment systems have shown immense potential over the past years, from everyday dietary monitoring to hospitalized setups \cite{papathanail2023nutritional,lu2020gofoodtm,ma2024mfp3d,ege2017estimating, zumthurm2025reducing}. 
However, a fundamental limitation of these tools lies in their inability to capture non-visual information, particularly differences in recipes and ingredient-level substitutions when it comes to home-cooked meals and obstructions within layered foods \cite{rahman2024food}. 
Composed meals often vary widely in their nutritional profiles depending on the specific ingredients used, which cannot be deduced from images alone and require further context. 
This poses a significant risk, particularly when dietary assessments inform critical decisions like insulin dosing recommendations, where inaccurate nutritional estimates may lead to serious health implications, such as severe hypoglycemic episodes  \cite{panagiotou2023complete}.

To overcome these limitations, dietary systems should adopt a holistic framework integrating recipes, along with contextual factors like user preferences, allergen sensitivities, cultural practices, and health goals. 
Traditional "one-size-fits-all" dietary guidelines often overlook these personal nuances, reducing their real-world effectiveness. 
Personalized nutrition bridges this gap by creating tailored plans while respecting individual needs, habits, and preferences like dietary restrictions (DR) with growing evidence highlighting positive impacts on health outcomes \cite{rouskas2025influence}.

More recently, there has been a surge in leveraging Large Language Models (LLMs) to power nutrition recommendation systems \cite{papastratis2024ai}. 
LLMs open up exciting possibilities for natural, conversational interaction with users, making it easier to ask questions and receive dietary guidance, a feature lacking from existing automatic dietary assessment systems.
However, LLMs are trained on vast, often unreliable, and culturally nonspecific internet data, making their recommendations sometimes untrustworthy and even potentially harmful \cite{niszczota2023credibility, ponzo2024chatgpt}.
Although LLMs have been applied to tasks like recipe generation \cite{kansaksiri2023smart}, many do not constrain outputs to verified knowledge or assess their credibility \cite{ruiz2024snapchef}, possibly leading to nutritionally unsound or harmful recommendations and eroding user trust \cite{ataguba2025exploring}.

Despite growing demand for nutritional awareness, nutritional information remains fragmented across sources. 
Diverse cultural, personal, and regional food practices call for flexible, unified guidelines that honor individual needs and local traditions. 
This underscores the need for a centralized, adaptable system to provide coherent, culturally sound, and personalized dietary recommendations. 
Building on these, we introduce the Swiss Food Knowledge Graph (SwissFKG), the first comprehensive and cohesive repository combining recipes, nutrient values sourced from the Swiss Food Composition Database (FCDB) \cite{nae_naehrwertdatenbanken}, allergen information, guidelines from the Swiss Food Pyramid (SFP) \cite{blv_ernaehrungsempfehlungen}, and DRs.
To enhance our graph, we propose a semi-automated enrichment pipeline leveraging LLMs. 
We further present a Graph-Retrieval Augmented Generation (Graph-RAG) pipeline tailored to question-answering (QA) over the SwissFKG. 
We evaluate LLMs and embedding models on various enrichment tasks, as well as their combined performance within the Graph-RAG framework over the SwissFKG.
Through this work, we lay the foundation for a general-population use case while paving the way toward a global Food Knowledge Graph (FKG) built on prior efforts \cite{gupta2024building,lei2021suggested,haussmann2019foodkg} to deliver verified, personalized nutrition-related recommendations that can be further integrated into existing automatic dietary systems.

\section{Related Works}

\subsection{Knowledge Graphs and Conversational Agents}

In a landscape increasingly shaped by LLMs, their growing adoption, driven by conversational abilities  \cite{sun2024building}, has led to widespread integration into applications with LLMs passing exams in multiple fields \cite{ZHOU2025101234,kataoka2023beyond}. 
Notably, chatbots \cite{huang2024does} are leveraging LLMs for QA tasks, while recommendation systems \cite{liu2024large} are incorporating them to allow users to express queries naturally. 
Notably, their use has expanded into critical domains such as healthcare, where conversational AI agents powered by LLMs assist in diagnosis and health guidance \cite{abbasian2023conversational}.
However, deploying LLMs for sensitive recommendations, particularly in healthcare, poses substantial risks if the underlying knowledge is not rigorously validated.
Usage of LLMs, to date, remains a popular topic of discourse due to their black-box nature \cite{ajwani2024llm,de2025unveiling} and the lack of credible audits of the web-scale data on which they are trained \cite{wang2025comprehensive}. 
These limitations leave LLMs prone to errors and hallucinations \cite{bang2025hallulens}, with LLMs often providing confident justifications for incorrect answers \cite{ajwani2024llm}.
To reduce such errors, techniques like Chain-of-Thought prompting (CoT) \cite{wei2022chain} were introduced to encourage step-by-step reasoning in LLMs, followed by Graph-RAG \cite{han2024retrieval} to ground responses in credible knowledge. 
Graph-RAG leverages well-constructed Knowledge Graphs (KGs) to retrieve verifiable information, thereby improving the credibility and transparency of LLM-based answers.

As such, there is an increasing interest in developing KGs that integrate disparate knowledge sources into unified repositories. 
KGs are graph-structured repositories representing entities as nodes and their relationships as edges \cite{hogan2021knowledge}, enabling the expression of rich, near-natural-language relationships. 
Generally, building a KG involves extracting and merging facts from various structured or unstructured sources, reconciling entities, and establishing relationships guided by an underlying ontology. 
KGs have found applications across numerous AI tasks, particularly in personalization, reasoning, and recommendation, notably within disease management \cite{SaraniRad2024}. 
In recommendation systems specifically, KGs serve as valuable sources of supplementary information, linking users  not only with the items they interact with, but also with the detailed attributes of those items. 
This facilitates more informed suggestions compared to collaborative filtering methods alone \cite{guo2020survey,raza2024comprehensive}. 
Consequently, KGs provide a structured foundation that allows AI systems to draw upon encoded domain knowledge, enabling more consistent and context-aware outcomes. 
Various methodologies for constructing KGs have been explored, ranging from fully manual approaches to semi-automated methods, where LLMs enrich KG data, to fully automated techniques in which LLMs autonomously define both the structure and content of the graph \cite{10.1145/3618295, kommineni2024humanexpertsmachinesllm,hu2025enablingllmknowledgeanalysis,zhang2025gkgllmunifiedframeworkgeneralized}.

\subsection{Food Knowledge Graphs}
In the food and nutrition domain, foundational ontologies and knowledge bases form the structured backbone for downstream applications. 
Early overviews document efforts to standardize food concepts and resolve vocabulary gaps across datasets \cite{vitali2018ons, lawrynowicz2022food, thornton2024reuse}. 
Among these, FoodOn classifies products, ingredients, and sources by organism, anatomical part, and processing method, enabling global traceability and quality control \cite{dooley2018foodon}.
Ontologies also underpin semantic recommendation systems, aligning food categories and nutrient relations with user profiles, yet they typically omit cooking steps and nuanced dietary preferences, leaving room for more recipe- and user-aware models \cite{mckensy2021ontology, padhiar2021semantic}.

Several projects build on this foundation to construct recipe-centric KGs. 
FoodKG integrates crowdsourced recipes with the USDA FCDB and maps ingredients to FoodOn classes, supporting nutritional QA and recommendations \cite{haussmann2019foodkg}. 
Likewise, Recipe1M+ has been used to link recipes, ingredients, and nutrient facts, but still treats each ingredient as a fixed node without modeling equivalents or substitutes \cite{marin2021recipe1m+, lei2021suggested}.
Others have even been more specific and have created national FKGs \cite{gupta2024building}. 
A parallel stream embeds nutritional and health considerations directly into FKGs. 
Diet-management ontologies encode food groups and nutrition targets, for instance, assisting diabetic menu planning \cite{min2022applications}.
RecipeKG goes further by computing healthiness scores against sugar-and-fat thresholds, enabling queries like “find a quick snack under 100 calories” and driving healthy recommendations \cite{recipekg2022}. 
However, these resources assume that meeting nutritional targets guarantees dietary appropriateness, neglecting critical factors such as food allergies, intolerance, and individual taste preferences.
In \cite{cordier2014taaable}, the authors use a case-based reasoning system with manually encoded substitution rules (e.g., butter→margarine for vegan variants) to tweak recipes to user needs. 
More recently, graph-embedding approaches rank substitutes by both co-occurrence similarity and a diet-aware substitution suitability metric that factors in allergen removal or calorie reduction \cite{shirai2021identifying}.

\section{Materials and Methods}

\subsection{Data collection}
\paragraph{Recipes}
We crawled and curated 1,000 recipes from well-known Swiss culinary websites. 
Along the way, we filtered out flawed entries with invalid ingredients, missing instructions, or exact duplicates. 
If multiple recipes shared a name but differed in content, we kept them all and assigned unique IDs. 
To guarantee structural consistency, we designed a custom JSON schema optimized for the KG construction and manually mapped all key entities.
Recipes gave a solid basis to our FKG by providing recipe macro-nutrition, a list of ingredients, utensils, instructions, seasons, cuisines, and certain keywords for tagging recipes.

\paragraph{Food databases}
To enhance the graph at the ingredient level, we enrich it with nutrient profiles for each ingredient.
For nutritional profiling, we primarily utilized the Swiss FCDB, which provides comprehensive macro- and micronutrient data for 1,146 food items commonly consumed in Switzerland \cite{nae_naehrwertdatenbanken}.
In cases where a specific ingredient was not available in the Swiss FCDB, the USDA's FoodData Central was used as a secondary reference source \cite{usda_fooddata_central_2019}. 
It features extensive, well-documented nutritional data for over 10,000 items, including branded and experimental foods.
To support ingredient-level substitutions, we incorporated data from Foodsubs \cite{foodsubs2025}, which lists substitution options for 3,177 unique ingredients.
We further included glycemic index (GI) values for ingredients in the KG to cater for the diabetic population.
Since GI values were missing from both the Swiss and USDA FCDBs, we sourced this data from FoodStruct \cite{foodstruct2025}, which compiles GI information from scientific publications and provides values for 615 individual food items or ingredients.

\paragraph{Guidelines and Recommendations}  
The integration of dietary recommendations is essential for a comprehensive FKG.
Switzerland defines 14 allergen categories as part of its food regulations, outlined in the regulations on Allergen Information \cite{blv_oidAl_2016}. 
These cover the mandatory labeling of common allergens like gluten, eggs, milk, nuts, and soy \cite{aha_food_allergy_2024}. 
However, the regulation only describes the categories; it does not provide a complete mapping from individual ingredients to their respective allergen category.  
We also wanted to include the SFP guidelines published by the Federal Food Safety and Veterinary Office \cite{blv_ernaehrungsempfehlungen}. 
These define nine main food groups according to the Swiss pyramid. While some of this information was already included in the Swiss FCDB, it wasn't available for all the ingredients we were working with.

\subsection{Data Enrichment}
\subsubsection{Translation of Non-English Content}   
Half of the recipes in our dataset were already in English, and for these, we also had French versions which we to evaluate translations. 
The rest were written in French, German, or Italian. We translated all non-English content (primarily French) using an LLM.  
To evaluate the LLM translation capabilities, we used the Crosslingual Optimized Metric for Evaluation of Translation \textbf({COMET})\cite{rei2020comet} score, an automatic metric that compares machine translations against human references. 
Unlike simpler methods, COMET considers meaning, fluency, and alignment with both the source and the reference text. 
We used the \texttt{wmt22-comet-da} model weights, focusing on ingredient lists and instructions, which are key components for FKG construction. 
COMET encodes three texts ($\mathbf{s}$, $\mathbf{t}$, $\mathbf{r}$) into vectors, combines them with their differences, and outputs a score correlating strongly with human evaluations. 
This ensures translations are both faithful to the source and fluent compared to references.

The COMET score is computed as follows:  

\begin{align}
\mathbf{h}_s &= \mathrm{Enc}(\mathbf{s}) \\
\mathbf{h}_t &= \mathrm{Enc}(\mathbf{t}) \\
\mathbf{h}_r &= \mathrm{Enc}(\mathbf{r})
\end{align}

\begin{equation}
\mathrm{COMET score}(\mathbf{s}, \mathbf{t}, \mathbf{r}) = 
\mathrm{FFN} \left( 
\left[
\mathbf{h}_s,\,
\mathbf{h}_t,\,
\mathbf{h}_r,\,
\mathbf{h}_s - \mathbf{h}_t,\,
\mathbf{h}_t - \mathbf{h}_r
\right]
\right)
\end{equation}


\noindent where
\[
\begin{aligned}
\mathbf{s} &\text{ is the original text,} \\
\mathbf{t} &\text{ is the LLM’s English translation,} \\
\mathbf{r} &\text{ is a reference translation,} \\
\mathrm{Enc}(\cdot) &\text{ converts text into vectors,} \\
\mathrm{FFN}(\cdot) &\text{ is the regression head that predicts} \\
                   &\quad \text{the quality of translation.}
\end{aligned}
\]

\subsubsection{Recipes: Cuisines, Diets, and Seasons' suitability}
To build rich entities and relationships, we used recipe names, descriptions, and keywords.
We applied a predefined list of the four seasons and 18 DR, based on WHO \cite{who_wpro_dietaryrestrictions_2021}, with religious diets further divided into multiple categories. 
We also used a list of 100 international cuisines to map recipes where possible. For example, \textit{“Vegan Swiss Summer Breadrolls”} implied Swiss cuisine, summer suitability, and a vegan (and therefore vegetarian) diet. 
All recipes were additionally labeled as “unrestricted” to indicate they are suitable for those without DR.

\subsubsection{Ingredients: Text Splitting, Allergens, SFP Categories, DRs}
\paragraph{Text Splitting} 
We found inconsistencies in the crawled recipe data, including irregular ingredient names, utensils listed as ingredients, non-standard terms, and misplaced or missing measurement units. 
To clean this, we used an LLM to normalize ingredient names to singular, canonical forms; extract preparation details into a notes field; correctly assign quantities and units; and move utensils to a separate field. 
For instance, ingredient text “a lemon, zest grated, and 1/2 juiced” becomes: Ingredient: lemon; Quantity: 1; Unit: –; Notes: zest grated; ½ juiced.
We then used a pattern matching approach based on \cite{ratcliff1988pattern} to score the output.
Given a predicted output string \(a\) and the expected string \(b\) - which we consider as ground truth, we define the similarity score as
\begin{equation}
\mathrm{Similarity}(a, b) \;=\; \frac{2M}{\lvert a\rvert + \lvert b\rvert}
\end{equation}

\noindent where
\[
\begin{aligned}
a   &\text{ is the predicted output string,}\\
b   &\text{ is the ground‐truth string,}\\
M   &\text{ is the total number of matching characters found by}\\
    &\quad\text{ recursively identifying longest common substrings.}\\
\lvert a\rvert &\text{ and }\lvert b\rvert\text{ denote the lengths of strings }a\text{ and }b\text{, respectively.}
\end{aligned}
\]
This score ranges from 0 (no overlap) to 1 (perfect match). By computing \(\mathrm{Similarity}(a,b)\), we obtain a single metric that reflects how closely each model’s output aligns with the expected split.

\paragraph{DR, Allergens, SFP Categories}
To enrich the KG with Swiss nutritional metadata, we used LLMs to infer DR to the 18 defined categories, assign allergens based on the 14 groups 
and map ingredients to the 9 SFP categories. 
Recipe-level DR were further enhanced based on the intersection of ingredient-level labels.
As an evaluation metric, we used F1 scores adapted to the label type.
For multi-label fields (e.g., allergen or pyramid categories), we computed a set-based F1 score using \cref{eq:set-f1}, while for binary diet flags, we used the standard F1-score per category in \cref{eq:binary-f1}, whereby the final diet score was computed as the mean F1 across all 18 DR labels.
\begin{equation}
F_1 =
\begin{cases}
1 & \text{if } S_\text{true} = S_\text{pred} = \emptyset \\[4pt]
0 & \text{if } |S_\text{true} \cap S_\text{pred}| = 0 \\[4pt]
\displaystyle\frac{2 \cdot P \cdot R}{P + R} & \text{otherwise}
\end{cases}
\label{eq:set-f1}
\end{equation}

\begin{equation}
P = \frac{|S_\text{true} \cap S_\text{pred}|}{|S_\text{pred}|}, \quad
R = \frac{|S_\text{true} \cap S_\text{pred}|}{|S_\text{true}|}
\label{eq:set-pr}
\end{equation}


\begin{equation}
F_1 = \frac{2 \cdot \text{Precision} \cdot \text{Recall}}{\text{Precision} + \text{Recall}}
\label{eq:binary-f1}
\end{equation}

\begin{equation}
\text{Precision} = \frac{TP}{TP + FP}, \quad
\text{Recall} = \frac{TP}{TP + FN}
\label{eq:binary-pr}
\end{equation}

where:
\[
\begin{aligned}
S_\text{true} &\ \text{is the set of true (ground-truth) labels},\\
S_\text{pred} &\ \text{is the set of predicted labels},\\
TP &\ \text{(true positives) = \# of labels correctly predicted},\\
FP &\ \text{(false positives) = \# of extra labels predicted},\\
FN &\ \text{(false negatives) = \# of true labels missed}.
\end{aligned}
\]

\subsubsection{Ingredient nutritional content and substitution}
While most recipes included energy and nutrient values, we extended this by attributing nutritional content to each ingredient (and substitute ingredient). 
We first queried the Swiss FCDB \cite{nae_naehrwertdatenbanken} and the USDA's FoodData Central \cite{usda_fooddata_central_2019} when needed using exact matches. 
If no direct match was found, we used embedding-based cosine similarity to identify semantically close ingredients, allowing us to recover information despite naming variations, synonyms, or minor spelling differences.
In our case, the cosine similarity measures how close two ingredients are in meaning based on their vector representations. 
Given two ingredient embeddings $\vec{\text{ing.}_1}$ and $\vec{\text{ing.}_2}$, the cosine similarity is computed as:

\begin{equation}
\text{Cosine Similarity}(\vec{\text{ing.}_1}, \vec{\text{ing.}_2}) = \frac{\vec{\text{ing.}_1} \cdot \vec{\text{ing.}_2}}{\|\vec{\text{ing.}_1}\| \|\vec{\text{ing.}_2}\|}  
\end{equation}

For evaluation, we framed the task as a retrieval problem and used a binary correctness criterion per query. 
For each query, we computed cosine similarities between its embedding and all candidate item embeddings, and selected the candidate with the highest similarity score as the predicted match. 
Although ties did not occur in our case, they can be handled by returning all top candidates and checking if the ground truth is among them, or resolving manually.
The accuracy can be given by:
\begin{equation}
\mathrm{Accuracy} \;=\; \frac{1}{N} \sum_{i=1}^N \mathbf{1}\bigl(X_i = Y_i\bigr)
\end{equation}

\noindent where
\[
\begin{aligned}
N   &\text{ is the total number of queries,}\\
X_i &\text{ is the predicted match for query }i\text{ (highest‐similarity item),}\\
Y_i &\text{ is the ground‐truth item for query }i,\\
\mathbf{1}(\cdot) &\text{ is the indicator function (1 if true, 0 otherwise).}
\end{aligned}
\]

\subsection{Knowledge Graph}
\begin{figure*}[ht]
    \centering
    \includegraphics[width=0.7\textwidth, height=0.5\textheight]{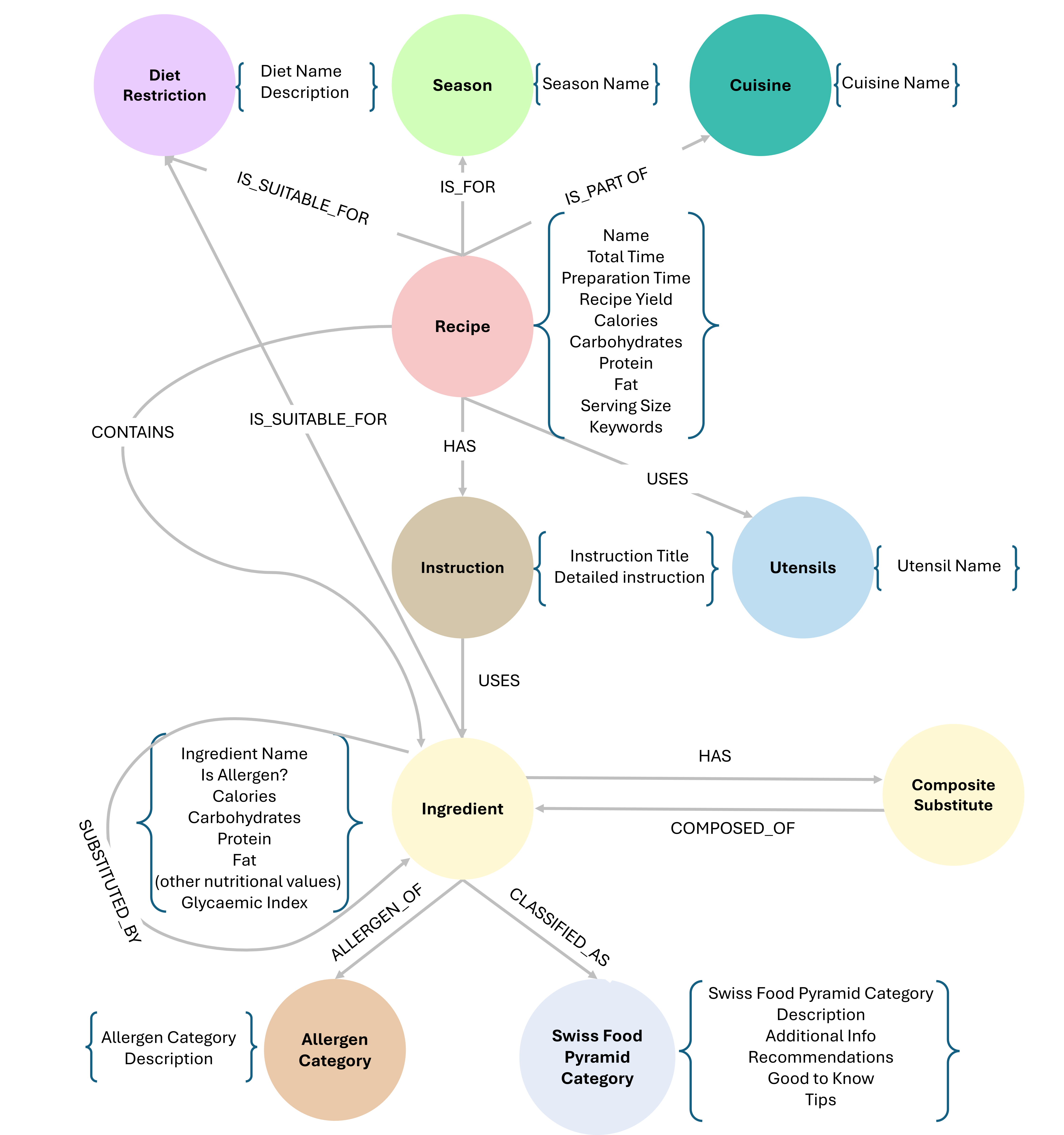}
    \caption{Knowledge Graph Nodes, Node Properties, and Relationships}
    \Description{Knowledge Graph Nodes, Node Properties, and Relationships}
\label{fig:onto_graph}
\end{figure*}
We built the SwissFKG based on the available data and according to the ontology shown in \cref{fig:onto_graph}. 
The KG includes the following node types: \textbf{Recipe}, \textbf{Instruction}, \textbf{Ingredient} (including \textit{substitute-ingredient} and \textit{composite substitute} variants), \textbf{DR}, \textbf{Season}, \textbf{Cuisine}, \textbf{Utensil}, \textbf{Allergen Category}, and \textbf{SFP Category}. 
Node properties are shown in the "\{\}".
The relationships connecting these entities are structured as follows:

\begin{itemize}
    \item \textbf{CONTAINS}: Connects a \textbf{Recipe} to its constituent \textbf{Ingredient}s, with associated quantity, unit, and preparation notes.

    \item \textbf{IS\_SUITABLE\_FOR}: Links a \textbf{Recipe} or an \textbf{Ingredient} to the \textbf{DR}s they comply with.

    \item \textbf{IS\_FOR\_SEASON}: Links a \textbf{Recipe} to a \textbf{Season} based on its contextual suitability.

    \item \textbf{IS\_PART\_OF}: Links a \textbf{Recipe} to a specific \textbf{Cuisine}.

    \item \textbf{USES}: Connects a \textbf{Recipe} to required \textbf{Utensil}s, with optional quantity and descriptive notes. Connects an \textbf{Instruction} to the \textbf{Ingredient}s used within that step, including quantity, unit, and notes.

    \item \textbf{HAS}: Connects a \textbf{Recipe} to its stepwise \textbf{Instruction}s.

    \item \textbf{ALLERGEN\_OF}: Connects an \textbf{Ingredient} to its associated \textbf{Allergen Category} (as defined in Swiss Annex~6).

    \item \textbf{CLASSIFIED\_AS}: Connects an \textbf{Ingredient} to a \textbf{SFP Category}, reflecting its nutritional classification.

    \item \textbf{SUBSTITUTED\_BY}: Links an \textbf{Ingredient} to a substitute \textbf{Ingredient}, along with the quantity ratio and any substitution notes.

    \item \textbf{HAS\_COMPOSITE\_SUBSTITUTE}: Links an \textbf{Ingredient} to a \textbf{Composite Substitute} node, which aggregates multiple ingredients.

    \item \textbf{COMPOSED\_OF}: Connects a \textbf{Composite Substitute} to its underlying \textbf{Ingredient}s, with associated substitution quantity and notes.
\end{itemize}

\subsection{Graph-RAG over SwissFKG}
To further demonstrate the practical applicability of the SwissFKG, we implemented a Graph-RAG pipeline \cite{han2024retrieval}. 
The knowledge in the graph is represented as structured triplets in the form (subject, predicate, object); for example, (Recipe:ApplePie, CONTAINS\_INGREDIENT, Ingredient:Apples), where each triplet expresses the factual relationship between two entities. 
We enrich this representation by incorporating both entity properties (e.g., the nutritional composition of apples) and relationship properties (e.g., the quantity of apples used in that relationship).
To enable semantic and structural retrieval, we generate graph embeddings \cite{liu2025large} that capture the meaning of entities, their relationships, and the broader graph topology. 
These embeddings allow us to efficiently retrieve relevant information based on user queries, supporting accurate and context-aware answer generation through the LLM.
When a user submits a query, our retrieval mechanism first uses an LLM to extract key concepts, keywords, and potential synonyms from the input. 
These extracted elements seed a search over graph nodes. 
In parallel, we embed the user query and use cosine similarity, with a cut-off of 0.5, against the graph embeddings to retrieve contextually aligned knowledge from the graph. 
We re-rank the retrieved knowledge and keep the 10 most relevant ones based on the cosine similarity to prevent exceeding the context length. 
By comparing the embeddings of the user query with those of the graph content, we identify and retrieve the most relevant knowledge from the graph. 
The final step involves passing this retrieved graph information to an LLM, which synthesizes the structured data into a coherent, natural-language answer. 
We further evaluated the performance of the Graph-RAG pipeline across different combinations of LLM and embedding models. 
For this, we used the following metric to compare the final output to the expected output:
\begin{equation}
\mathrm{Accuracy} \;=\; \frac{1}{N} \sum_{i=1}^N \mathbf{1}\left( Y_i \subseteq X_i \right)
\end{equation}

\noindent where
\[
\begin{aligned}
N &\text{ is the total number of questions evaluated,} \\
X_i &\text{ is the model's response to question } i, \\
Y_i &\text{ is the expected answer to question } i, \\
\mathbf{1}(\cdot) &\text{ is the indicator function, which returns 1 if the} \\
                  &\quad \text{condition } Y_i \subseteq X_i \text{ is true, and 0 otherwise.}
\end{aligned}
\]

\subsection{Experimental Setup}
All experiments were conducted on NVIDIA RTX A6000 GPU, with LLMs and embedding models served through Ollama (version 0.9.0) \cite{ollama_2025}. 
To ensure a fair comparison, we used the same prompt and fixed generation settings across all LLMs: zero temperature for deterministic output, a fixed random seed for reproducibility, a 4096-token context window, and restrictive sampling (top-$p=0$, top-$k=1$). 
For models that support internal reasoning, the \textit{thinking} mode was explicitly disabled. 
For LLM system prompts, the CoT prompting technique was used.
Additionally, for all enrichment tasks, LLMs were asked to respect a specific JSON format when providing outputs with examples provided as part of system prompts demonstrating sample input and expected output. 

To address incomplete or erroneous data in our initial knowledge sources and prepare them for integration into the FKG, we leveraged LLMs to enrich content according to our target ontology. 
To balance performance, multilingual support, and efficiency under limited resources, we selected recent (2024–2025) open-source LLMs under 70B parameters.
These models performed well on benchmarks, especially instruction-following, and offered a strong trade-off between capability and cost for the enrichment task. We used the following models: \texttt{Phi-4 (14B)} \cite{microsoft_phi4_2024}, \texttt{Mistral Small 3.2 (24B)} \cite{mistral_small3.2_2506}, \texttt{Gemma3 (27B)} \cite{deepmind_gemma3_2025}, and \texttt{Qwen3  (30B-A3B)} \cite{qwen3_2025}.
For identifying semantically similar nutrition terms, suggesting ingredient substitutions, and mapping GI values, we experimented with the following embedding models: texttt{All MiniLM 33M} \cite{reimers_all-MiniLM-L6-v2_2020},\texttt{Nomic Embed} \cite{nussbaum2024nomic_embed_text_v1},  \texttt{Mxbai Embed Large} \cite{li2023angle}. 


We evaluated the LLM and embedding models on various enrichment tasks using a curated expected output considered as ground truth covering ~10\% of the data: 100 recipes and 200 ingredients. 
For translation quality, we extracted English–French recipes and had a native bilingual speaker verify translations across three semantic targets: name, instructions, and ingredients. 
For ingredient-related tasks, including text splitting, mapping DR, allergens, SFP categories, and ingredient matching, we constructed dedicated ground truths for the ingredient entries.
To demonstrate SwissFKG’s applicability as a proof of concept, we combined the two best-performing LLM models from the enrichment tasks with the different embedding models in a Graph-RAG pipeline. 
We evaluate this setup on 50 curated questions generated from the KG to measure the effectiveness of prompting LLMs within our approach.

\section{Results}
\subsection{Data Enrichment}
\begin{figure}[ht]
    \centering
    \includegraphics[width=0.5\textwidth]{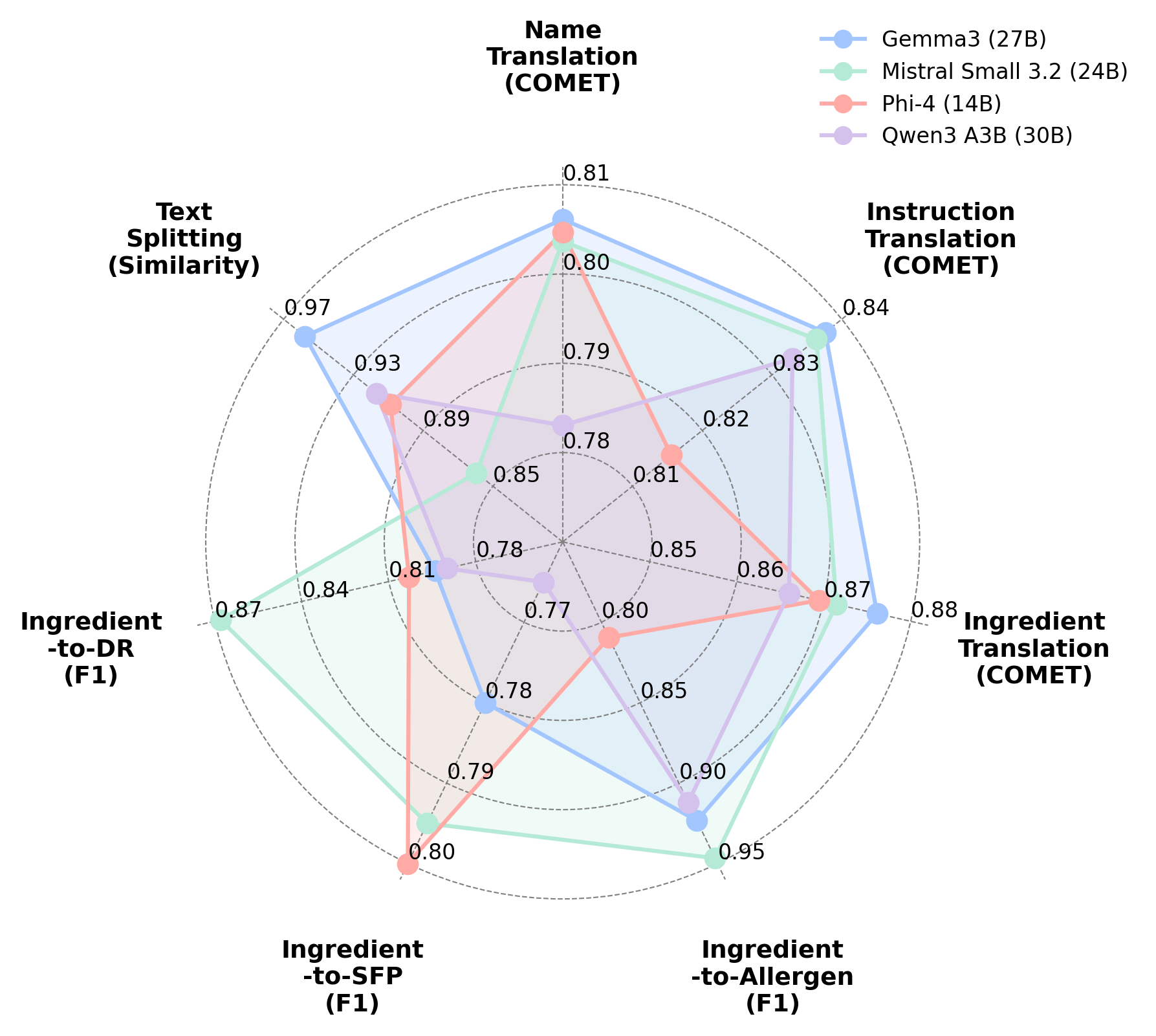}
    \caption{Performance of LLMs on enrichment tasks. COMET scores reported for translation tasks, Similarity for text splitting task and F1 scores for  Ingredient-to-allergen, -SFP, and -DR mappings}
    \Description{Performance of LLMs on enrichment tasks. COMET scores reported for translation tasks, Similarity for text splitting task and F1 scores for  Ingredient-to-allergen, -SFP, and -DR mappings}
\label{fig:spider_plt}
\end{figure}

\subsubsection{Translation of Non-English Content}
Table~\ref{tab:comet_scores} reports average COMET scores per model. 
\textbf{Gemma3 (27B)} consistently outperformed others in all aspect, with scores of \textbf{\underline{0.8061}}, \textbf{\underline{0.8376}}, and \textbf{\underline{0.8761}} indicating strong alignment with reference translations in food-specific contexts.
\textit{Mistral Small 3.2 (24B)} followed closely, particularly on instructions and ingredients.
\textit{Phi-4 (14B)} performed reasonably but dropped on instructions. 
Notably, \textit{Qwen3  (30B-A3B)} yielded the lowest scores overall, though it still maintained reliability with all scores above 0.78.

\begin{table}[!ht]
\centering
\small
\begin{tabular}{lccc}
\toprule
\textbf{Model} & \textbf{Name} & \textbf{Instructions} & \textbf{Ingredients} \\
\midrule
Gemma3 (27B)            & \textbf{\underline{0.8061}} & \textbf{\underline{0.8376}} & \textbf{\underline{0.8761}} \\
Mistral Small 3.2 (24B)  & 0.8037 & 0.8364 & 0.8715 \\
Phi-4 (14B)              & 0.8047 & 0.8156 & 0.8694 \\
Qwen3  (30B-A3B)         & 0.7831 & 0.8329 & 0.8660 \\
\bottomrule
\end{tabular}
\caption{COMET scores for translation tasks}
\label{tab:comet_scores}
\end{table}

\subsubsection{Ingredients: Text Splitting, Allergens, SFP Categories, DRs}
\paragraph{Text Splitting}
Table~\ref{tab:similarity_scores} gives an insight to how well the models were in following the given instructions for splitting text and following examples provided to them.
\textbf{Gemma3 (27B)} achieved the highest similarity score of \textbf{\underline{0.9578}}.
\textit{Phi-4 (14B)} followed closely with scores above 0.91, while \textit{Qwen3  (30B-A3B)} also performed reliably.
\textit{Mistral Small 3.2 (24B)} showed weaker performance here. 

\begin{table}[!ht]
\centering
\small
\begin{tabular}{lc}
\toprule
\textbf{Model} & \textbf{Text Splitting} \\
\midrule
Gemma3 (27B)            & \textbf{\underline{0.9578}} \\
Mistral Small 3.2 (24B)  & 0.8628 \\
Phi-4 (14B)             & 0.9197 \\
Qwen3  (30B-A3B)         & 0.9178 \\
\bottomrule
\end{tabular}
\caption{Similarity scores for ingredient text splitting}
\label{tab:similarity_scores}
\end{table}

\paragraph{DR, Allergens, SFP Categories}
As shown in Table~\ref{tab:f1_scores}, \textbf{Mistral Small 3.2 (24B)} achieved the best results in both allergen and DR mappings with an F1 score of \textbf{\underline{0.947}} and \textbf{\underline{0.868}} respectively, demonstrating good generalization capabilities for the nutrition domain. 
For SFP mapping, \textit{Phi-4 (14B)} performed best with an F1 score of \textbf{\underline{0.8}}, despite its smaller size. 
\textit{Gemma3 (27B)} and \textit{Qwen3  (30B-A3B)} performed generally well for allergen detection, while for SFP Categorization and DRs, results among LLMs were very close.
The majority of errors occurred when the mappings should have been none, i.e., no category should have been attributed. 
For example, with ingredient "icing sugar", the expected allergen should have been None - yet all LLMs except \textbf{Mistral Small 3.2 (24B)} assumed an allergen of category 1: gluten-containing products. 
Looking at DR mappings across LLMs, the diabetic diet had the highest error rates, ranging from 61-74\%. 
On the other hand, gluten-free and some religious diets were among the easiest for LLMs to map, with error rates as low as 7\%.

\begin{table}[!ht]
\centering
\small
\begin{tabular}{lccc}
\toprule
\textbf{Model} & \textbf{Allergen} & \textbf{SFP} & \textbf{DR} \\
\midrule
Gemma3 (27B)            & 0.923 & 0.78 & 0.794 \\
Mistral Small 3.2 (24B)  & \textbf{\underline{0.947}} & 0.795& \textbf{\underline{0.868}} \\
Phi-4 (14B)              & 0.810 & \textbf{\underline{0.8}} & 0.803 \\
Qwen3  (30B-A3B)         & 0.912 & 0.765 & 0.790 \\
\bottomrule
\end{tabular}
\caption{F1 scores for Ingredient-to -Allergen, -SFP, and -DR mapping tasks}
\label{tab:f1_scores}
\end{table}

\subsubsection{Ingredient nutritional content and substitution}
Using cosine similarity as the retrieval criterion, we found that, as shown in Table~\ref{tab:cosines_matching}, \textbf{Mxbai Embed Large} achieved the highest accuracy at \textbf{\underline{0.66}}, slightly outperforming \textit{All Minilm 33M}, followed by \textit{Nomic Embed}. 
Looking deeper, ingredient matches using \textit{All MiniLM 33M} often resulted in low cosine similarity scores (<0.5), despite correct alignments.
For example, the ingredient "Nori" was correctly matched with "Seaweed, Nori, dried" with a similarity of 0.47, while \textit{Nomic Embed} yielded a higher score of 0.78 for the same pair.

\begin{table}[!ht]
\centering
\begin{tabular}{lc}
\hline
\textbf{Model} & \textbf{Accuracy} \\
\hline
All Minilm 33M & 0.64 \\
Mxbai Embed Large & \textbf{\underline{0.66}} \\
Nomic Embed & 0.61 \\

\hline
\end{tabular}
\caption{Accuracy of closest ingredient matching}
\label{tab:cosines_matching}
\end{table}

\subsection{Knowledge Graph}
To construct the KG, we leveraged the top-performing models from prior enrichment tasks - Gemma3 (27B) \cite{deepmind_gemma3_2025} for translation tasks and text splitting, Mistral Small 3.2 (24B) \cite{mistral_small3.2_2506} for allergens and DRs mappings, Phi-4 (14B) \cite{microsoft_phi4_2024} for SFP Category mappings, and ingredient name matching was done using Mxbai Embed Large \cite{li2023angle}. 
All the data together resulted in a SwissFKG with 5,896 nodes and 62,499 relations. 
As shown in Table~\ref{tab:node_dist}, the node types with the largest counts were \textit{Ingredient (including substitutions)} (2,548), \textit{Instruction} (2,176), and \textit{Recipe} (1,000).
The breakdown of the relationships of the graph is given in Table~\ref{tab:relationship_counts}, whereby we noted a low number of recipes and cuisines being linked, highlighting missing cuisine tagging in a lot of instances from the crawled data. 
Our SwissFKG comprises 893 recipes that contain ingredients that are allergens, while the rest are without.
Additionally, the top 5 most used ingredients were salt, pepper, olive oil, pepper, water, and sugar. 
Allergens that appeared the most were of categories 7 (Dairy/Milk), 1 (Glutens), and 8 (Nuts). 
Among the 1000 recipes, 584 are vegetarian friendly, 467 gluten-free, over 500 were suitable for religious restrictions, and above 300 were low/free of lactose.  

\begin{table}[h!]
\centering
\small
\begin{tabular}{lc}
\toprule
\textbf{Label} & \textbf{Count} \\
\midrule
Ingredient                & 2548 \\
Instruction               & 2176 \\
Recipe                    & 1000 \\
Utensil                   & 91 \\
Cuisine                   & 21 \\
DietRestriction           & 19 \\
AllergenCategory          & 14 \\
CompositeSubstitute       & 14 \\
SwissFoodPyramidCategory  & 9 \\
Season                    & 4 \\
\bottomrule
\end{tabular}
\caption{Entity label distribution in the annotated dataset.}
\label{tab:node_dist}
\end{table}

\begin{table}[ht]
  \centering
  \resizebox{0.5\textwidth}{!}{
  \begin{tabular}{lll r}
    \toprule
    Source Labels               & Relationship Type    & Target Labels              & Count   \\
    \midrule
    Ingredient           & IS\_SUITABLE\_FOR    & DietRestriction       & 33694   \\
    Recipe                & CONTAINS             & Ingredient            & 10099   \\
    Instruction           & USES     & Ingredient            & 6073    \\
    Recipe                & IS\_SUITABLE\_FOR    & DietRestriction       & 6008    \\
    Ingredient            & CLASSIFIED\_AS       & SwissFoodPyramidCategory      & 2301    \\
    Recipe                & HAS     & Instruction           & 2176    \\
    Ingredient            & ALLERGEN\_OF         & AllergenCategory      & 1037    \\
    Recipe                & USES        & Utensil               & 548     \\
    Recipe                & IS\_FOR      & Season                & 271     \\
    Ingredient            & SUBSTITUTED\_BY      & Ingredient            & 158     \\
    Recipe                & IS\_PART\_OF         & Cuisine               & 67      \\
    CompositeSubstitute    & COMPOSED\_OF         & Ingredient            & 46      \\
    Ingredient            & HAS    & CompositeSubstitute    & 21      \\
    \bottomrule
  \end{tabular}
  }
  \caption{Counts of relationship types between node labels}
  \label{tab:relationship_counts}
\end{table}

\subsection{Graph-RAG over SwissFKG}
Table \ref{tab:embedding_accuracy} presents the performance of the two best overall performing  LLMs (based on enrichment tasks) when integrated into the Graph-RAG pipeline using different embedding models over the SwissFKG dataset. 
Among the tested LLMs, \textbf{Gemma3 (27B)} achieved the highest accuracy of \textbf{0.80} when paired with the \textbf{Mxbai Embed Large} embedding model.
Both Gemma3 (27B) and Mistral Small 3.2 (24B) performed the best using the Mxbai Embed Large embedding model, and the worst with All MiniLM 33M. 
Swapping embedding models with the same LLM led to up to a 16\% difference in accuracy for Gemma3 (27B) and 10\% for Mistral Small 3.2 (24B).
Furthermore, we note that with the All MiniLM 33M embedding model, no knowledge could be retrieved for 8 questions with Gemma3 (27B) and 13 questions with Mistral Small 3.2 (24B).
\label{subs:GR_Ove_SWS}
\begin{table}[ht]
\centering
\begin{tabular}{llc}
\hline
\textbf{Model Name} & \textbf{Embedding Model} & \textbf{Accuracy} \\
\hline
Gemma3 (27B) & All MiniLM 33M              & 0.64 \\
Gemma3 (27B) & Mxbai Embed Large     & \textbf{\underline{0.80}} \\
Gemma3 (27B) & Nomic Embed      & 0.72 \\
Mistral Small 3.2 (24B) & All MiniLM 33M      & 0.66 \\
Mistral Small 3.2 (24B) & Mxbai Embed Large & 0.76 \\
Mistral Small 3.2 (24B) & Nomic Embed  & 0.74 \\
\hline
\end{tabular}
\caption{Accuracy of QA through Graph-RAG over SwissFKG}
\label{tab:embedding_accuracy}
\end{table}

\section{Discussion}
Our comprehensive evaluation of LLM-based data enrichment and Graph-RAG over the SwissFKG reveals several key insights into the interplay between model scale, pretraining alignment, retrieval representations, and domain-specific performance.
Contrary to common assumptions that larger parameter counts inherently yield superior performance, our results demonstrate that model scale is not the sole determinant of success across enrichment tasks. 
As \cref{fig:spider_plt} illustrates, Gemma3 (27B) achieved the highest COMET scores on translation of names, instructions, and ingredients (Table \ref{tab:comet_scores}), reflecting the strength of its multilingual pretraining. 
Mistral Small 3.2 (24B), despite being smaller than Gemma3 (27B), outperformed all others on allergen mapping and DR classification. 
Phi-4 (14B), even though being the smallest model, achieved the best alignment with SFP categories.
These insights underscore the fact that certain models are more apt for certain tasks, and following instructions, irrespective of their size.
Despite explicit instructions to avoid unstated assumptions, LLMs made certain assumptions in allergen and SFP mappings.
That misstep underscores how some LLMs might fail to internalize task directives and might not follow given examples.
Even when All MiniLM correctly matches ingredients, it often assigns low similarity scores, likely due to tightly clustered embeddings that obscure true matches,suggesting that fine-tuning on food-specific data, rather than using a fixed cutoff like 0.5, could improve performance.

The poor performance of most LLMs in tagging the diabetic DR likely stems from clinical guidelines favoring personalized patterns, nutrient goals, and portion control over fixed lists, making the “diabetic diet” highly variable, and from the sparse or conflicting examples in general web-trained models due to the lack of standardized guidelines \cite{sinska2023dietary}.
The low number of relationships linked recipes to cuisines flags a key area for improvement within the graph.
Beyond the enrichment stage, Graph-RAG experiments over our SwissFKG, even though with promising results of 80\% over 50 questions, highlights the critical role of embedding representations in retrieval-augmented inference.
The lower performance particularly in the zero-information retrieval incidents (\cref{subs:GR_Ove_SWS}), further underscores the need for careful embedding selection when constructing Graph-RAG pipelines relying on graph embeddings.

While our approach yielded promising results, several limitations merit attention.
Model size remains a constraint, as we evaluated only non-domain-specific off-the-shelf LLMs up to 30B parameters; larger or domain-adapted models may further enhance enrichment quality. 
Our choice of metrics, COMET, character similarity, F1, and cosine accuracy, captures only part of the picture, overlooking hallucination rates, semantic drift, human alignment, and probable structural inconsistencies with the graph schema. 
The current retrieval strategy, relying on static embeddings and cosine similarity, could be enhanced by incorporating hybrid methods \cite{mandikal2024sparsemeetsdensehybrid, zhang2024efficient} that combine dense semantic and sparse lexical retrieval or adaptive techniques \cite{taguchi2025efficient, hsu2025dat} that dynamically adjust thresholds and retrieval based on query context, to improve relevance and robustness.
Finally, automated enrichment, while efficient, still requires expert oversight for low-confidence or novel mappings to safeguard data integrity.

As future works, we propose to further expand the model repertoire and possibly fine-tune models to be more domain-specific, particularly embedding models, deploying a dynamic ingestion pipeline to keep nutritional data current, and introducing a human-in-the-loop interface for validating uncertain graph entries. 
Furthermore, extending the Graph-RAG pipeline with multimodal inputs and integrating multimodal LLMs for QA could enable processing of meal photographs and evolve into a more capable nutritional agent.
User-centered evaluation with realistic professional queries could help refine and stress-test the system for reliable deployment in real-world settings.
A promising direction for the future is integrating the SwissFKG into personalized dietary systems by combining QA over the knowledge along with automatic dietary assessment systems. 
Such integration could enable real-time, context-aware nutritional guidance, thus bridging the gap between static food databases and dynamic individual consumption patterns. 
As the need for FKGs grows, our work contributes to ongoing efforts to standardize and harmonize FKGs and can serve as a foundation for assembling these into a unified global FKG, fostering cross-border collaboration in nutrition research.
While this work showcases a general-use case for public nutrition recommendations, we aim to extend the graph’s applications to advanced domains like clinical nutrition, food chemistry, sustainability, and functional food research.

\section{Conclusion}
In this work, we introduced SwissFKG, the first FKG, to the best of our knowledge, that integrates Swiss recipes, ingredients, nutritional profiles, allergen classifications, DR, and national nutrition guidelines into a unified repository.
At the core of SwissFKG is a semi-automated enrichment pipeline powered by low-resource (<70B parameters), open-source LLMs, demonstrating satisfactory results across enrichment tasks.
We further demonstrated the practical value of SwissFKG by using it as the backbone for a Graph-RAG pipeline that enables LLMs to answer user queries grounded in structured food knowledge with up to 80\% accuracy over 50 questions.
Our work lays an important foundation for future research towards building a global FKG and paves the way for integration with personalized dietary systems, where nutritional insights from SwissFKG could enhance automatic dietary assessment systems for personalized precision nutrition.






\begin{acks}
This work was partly supported by the European Commission and the Swiss Confederation - State Secretariat for Education, Research and Innovation (SERI) within the projects 101057730 Mobile Artificial Intelligence Solution for Diabetes Adaptive Care (MELISSA) and 101080117 Preventing Obesity through Biologically and Behaviorally Tailored Interventions for You (BETTER4U).
\end{acks}

\clearpage

\bibliographystyle{ACM-Reference-Format}
\bibliography{sample-base}


\end{document}